# Addressing accuracy and hallucination of LLMs in Alzheimer's disease research through knowledge graphs

Tingxuan Xu[1]    Jiarui Feng[1]    Justin Melendez[1]    Kaleigh Roberts[1]
Donghong Cai[1]    Mingfang Zhu[1]    Donald Elbert[2]
Yixin Chen[1*]    Randall J. Bateman[1*]

[1]Washington University in St. Louis    [2]University of Washington

* Co-senior authors

## Abstract

In the past two years, large language model (LLM)-based chatbots, such as ChatGPT, have revolutionized various domains by enabling diverse task completion and question-answering capabilities. However, their application in scientific research remains constrained by challenges such as hallucinations, limited domain-specific knowledge, and lack of explainability or traceability for the response. Graph-based Retrieval-Augmented Generation (GraphRAG) has emerged as a promising approach to improving chatbot reliability by integrating domain-specific contextual information before response generation, addressing some limitations of standard LLMs. Despite its potential, there are only limited studies that evaluate GraphRAG on specific domains that require intensive knowledge, like Alzheimer's disease or other biomedical domains. In this paper, we assess the quality and traceability of two popular GraphRAG systems. We compile a database of 50 papers and 70 expert questions related to Alzheimer's disease, construct a GraphRAG knowledge base, and employ GPT-4o as the LLM for answering queries. We then compare the quality of responses generated by GraphRAG with those from a standard GPT-4o model. Additionally, we discuss and evaluate the traceability of several Retrieval-Augmented Generation (RAG) and GraphRAG systems. Finally, we provide an easy-to-use interface with a pre-built Alzheimer's disease database for researchers to test the performance of both standard RAG and GraphRAG.

## 1   Introduction

The development of large language models (LLMs) represents a major milestone in artificial intelligence, with transformative impacts across diverse domains. LLMs, such as OpenAI's GPT models[1] and Anthropic's Claude[2], demonstrate an unprecedented ability to generate human-like responses, synthesize information, and adapt to a wide variety of tasks. These capabilities have driven innovations in industries ranging from education and e-commerce to public services and healthcare.

Despite their impressive capabilities, current LLMs face major limitations that preclude their reliable use in biomedical science, for several reasons. First, these models often underperform in specialized contexts because their training primarily relies on general knowledge[3]. This gap in expertise hinders their ability to provide precise and contextually accurate answers to complex scientific questions[4]. Second, the explainability and reliability of AI systems are critical in scientific research, as researchers depend on verifiable information to draw conclusions or guide further inquiry. Of particular importance is the ability to accurately reference prior information through appropriate citations. Ideally, this explainability can be ensured by providing well-cited, relevant references alongside the answers. However, LLMs frequently suffer from hallucinations, producing non-existent, irrelevant, or fabricated references that compromise their trustworthiness[5]. These limitations not only reduce the applicability of LLMs in scientific workflows but also pose challenges to their integration into rigorous academic and professional environments. Recently, Retrieval-Augmented Generation (RAG)[6] particularly its graph-based variant, GraphRAG[7,8] has gained significant attention for its ability to enhance LLM-based chatbots by integrating large-scale, domain-specific databases. RAG systems operate by first indexing domain-specific knowledge, such as scientific papers, into a structured database optimized for efficient querying. When presented with a question, the system retrieves relevant knowledge from this database and combines it with the query. The LLM then processes both the query and the retrieved information to generate a response. This approach enables the LLM to review contextually relevant information, improving the reliability of generated answers and mitigating issues such as hallucinations. GraphRAG builds on the foundational principles of RAG but employs a graph-based structure for the knowledge base[9]. This graph representation offers a more organized framework for storing large-scale knowledge, which enhances retrieval performance and facilitates more effective thinking.

GraphRAG shows great promise as a solution for scientific chatbots by enhancing domain-specific knowledge integration for general LLMs while providing traceability for generated answers. However, there is currently only limited evaluation of GraphRAG's potential in biomedical domains in the existing literature[10]. In this paper, we take the first step toward assessing the performance and traceability of GraphRAG to answer scientific questions related to Alzheimer's disease. Specifically, we focus on two widely recognized GraphRAG systems: Microsoft GraphRAG[8] and LightRAG[7]. To conduct our evaluation, we constructed a knowledge base comprising 50 research papers on Alzheimer's disease that are coauthored by senior authors in our group. We then collected 70 questions, curated by professional researchers in the field, to assess the capabilities of these systems. Our evaluation focuses on five key metrics: accuracy, comprehensiveness, diversity, empowerment, and directness of the answers generated by GraphRAG. We compare these results with those of general-purpose chatbots to understand the relative advantages of GraphRAG. Additionally, we analyze the ability of the two GraphRAG systems to provide accurate and traceable references alongside their answers. We find that the current GraphRAG systems are still limited in providing detailed references. This lack of transparency obscures traceability and hinders reproducibility, especially in fields such as biomedical research, where the ability to verify conclusions against specific sources is critical. Finally, we construct an easy-to-use human interface with pre-built Alzheimer's disease database for researchers to test and evaluate the performance of both standard RAGs and GraphRAGs.

In summary, our contributions are as follows:

- We systematically evaluate the performance of two widely used GraphRAG systems in answering scientific questions that require specialized domain knowledge.

- We provide an in-depth analysis of the systems' ability to generate accurate and reliable context and references alongside their answers.

- We construct an easy-to-use human interface for researchers in the field of Alzheimer's disease to explore the potential of GraphRAGs.

These contributions aim to evaluate and advance the capabilities of the GraphRAG framework by addressing key challenges in accuracy and traceability, thereby demonstrating its practical utility for scientific research applications.

## 2    Related Works

Since the release of ChatGPT[1], the concept of large language models (LLMs) - based AI assistants has become popular, due to their superior ability to handle various tasks through their reasoning ability using natural language. Since then, there has been abundant research aimed at adapting LLMs to different fields, including biomedical fields. Despite the current success of LLMs, there are still many limitations that hinder their application in domain-specific scenarios. First, the training of the LLMs is often resource and time costly, which makes it hard to include up-to-date knowledge. Further, most of the training corpus is from general internet text, which limits its utility in a knowledge-intensive domains. Second, current LLMs can only handle input text of a specific length[11]. This makes it hard for them to reason over long input, like documents or books. Despite recent efforts to extend the context windows of LLMs[12,13], long-context reasoning  still faces challenges like being lost in the middle[14]. Third, hallucination is a long-standing problem for many state-of-the-art LLMs. Namely, the LLMs tend to output grammatically correct but semantically meaningless or even wrong information.  One particular example is that when a user asks the LLMs to provide a reference for their answer, the LLMs often provide confabulated references to non-existent publications [5].

One of the root causes for the above problems is that LLMs save the knowledge of the world using its billion parameters, which is implicit and complicated. To overcome it, one solution is to provide LLMs with an external knowledge base that is accurate, updated, and comprehensive, such that LLMs can leverage this information for better task-solving ability. That's the basic principle of the retrieval-augmented generation

(RAG)[15]. A basic RAG system consists of three components, a database, a retriever, and an LLM. Given a query from the user, the RAG first uses the retriever to retrieve relevant knowledge or information from the database. Then, the retrieved information will be used as context for LLM to answer the query from the user. After RAG was proposed, there have been many efforts that try to optimize the database and retriever. One successful variant is called GraphRAG, where the database is a large-scale knowledge graph and the retriever can retrieve information from the graph. Some famous frameworks include Microsoft GraphRAG[8], LightRAG[7], HippoRAG[16], and SubgraphRAG[17]. PaperQA[18] proposed an agentic RAG system, which can retrieve relevant information from websites, such as searching PubMed for relevant publications. More recently, MedGraphRAG[10] applied GraphRAG medicine, an application that shows great potential over standard RAG systems. However, there is still limited understanding of how well GraphRAG performs in the biomedical domain versus standard LLMs.

There are also task-specific RAG models. A representative example is AlzheimerRAG[21]. AlzheimerRAG is a multimodal system for Alzheimer's research that combines text and image information through fine-tuned models and cross-modal attention fusion. It processes Alzheimer's case studies, retrieves domain-specific clinical text, and integrates image features using LLaVA. This design highlights the use of multimodal integration for domain-specific accuracy, in contrast to the general frameworks described in the following sections.

## 3   Overview of RAG and GraphRAG

In this section, we first briefly explain how RAG works in general and then dive into both Microsoft GraphRAG and LightRAG respectively. In Figure 1, we show the workflow of both Microsoft GraphRAG and LightRAG.

### 3.1   A general framework of RAG

A typical RAG system consists of three essential components: indexing , retrieval , and answering . 1) The indexing component is responsible for converting an external database into a specific structural database. 2) The retrieval component is used to fetch relevant information from the structural database based on the input query. 3) The answering component takes both the retrieved information and the query as input to generate responses. Typically, the answering component is implemented by a powerful LLM-based chatbot. Formally, let a RAG system be denoted as $S$, a general RAG framework can be defined as follows:

$$S = \left(A, R, \widehat{D}\right),$$
$$S(q; D) = A\left(q, \delta\left(q; \widehat{D}\right)\right),$$
$$\widehat{D} = \varphi(D).$$

The $A$ and $R$ represent the answering module and retrieval module. The $\varphi(\cdot)$ is the indexing function, which converts an external data source $D$ into structural databases $\widehat{D}$ that is easy for retrieval. the $\delta(\cdot)$ is the retrieval function usually implemented by comparing the query against the indexed data. Given a query $q$, the RAG system S first uses retrieval function $\delta(\cdot)$ to retrieve relevant information from $\hat{D}$ and that feeds it into answering module $A$ to generate an answer.

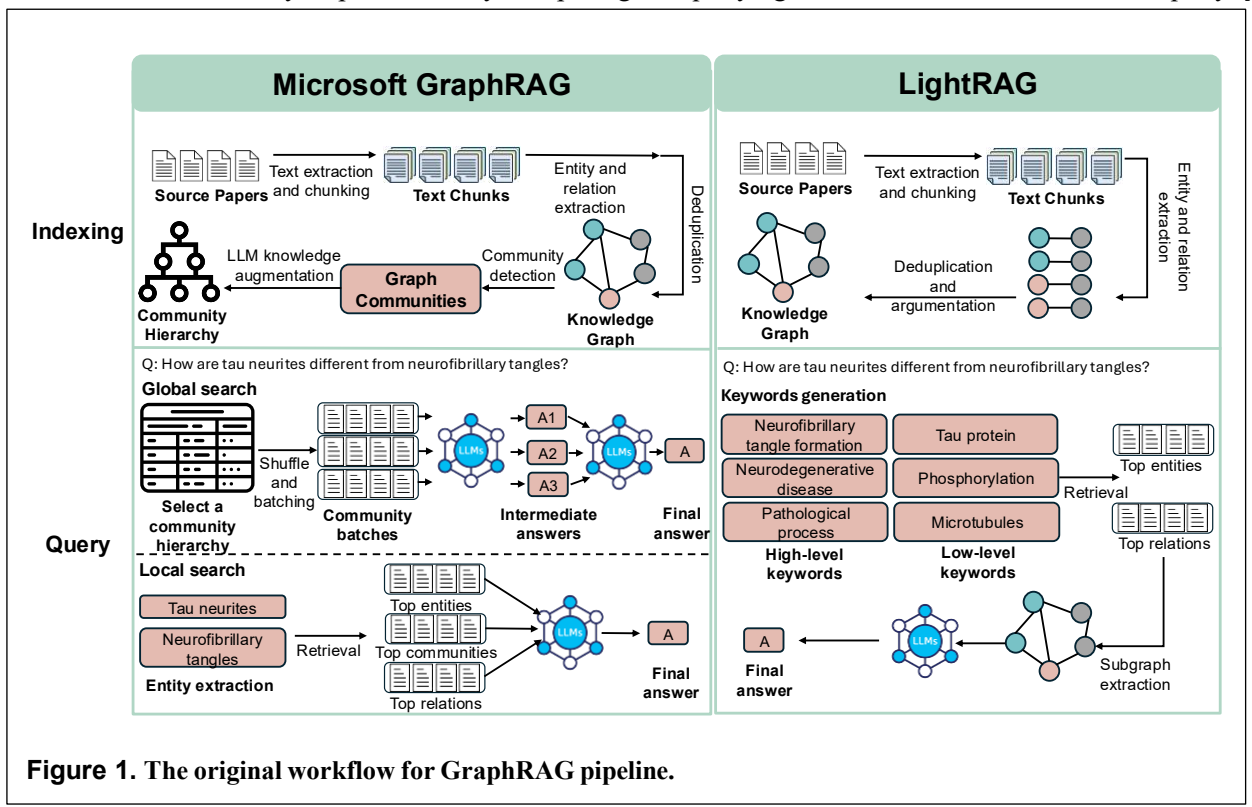

**Figure 1.** The original workflow for GraphRAG pipeline.

## 3.2 Microsoft GraphRAG

The GraphRAG framework, developed by Microsoft, integrates graph-based reasoning with retrieval-augmented generation (RAG) to provide query-focused answers derived from large document collections. In the left panel of Figure 1, we show the detailed workflow of the Microsoft GraphRAG. In the following, we briefly describe the three components implemented in Microsoft GraphRAG.

**Indexing.** Unlike the standard RAG system, Microsoft GraphRAG introduces a sophisticated indexing mechanism that involves constructing large-scale knowledge graphs and hierarchical community structures to enhance downstream retrieval. Specifically, the process begins with dividing raw input documents into smaller text units. For each text unit, the system extracts entities and relationships, forming graph components. These components are then combined and deduplicated to generate a comprehensive knowledge graph representing the input documents. To enrich the graph, LLMs are employed to augment the extracted entities and relationships with additional contextual information. Subsequently, the system applies the Leiden algorithm to build a hierarchical community structure within the graph. This structure partitions the graph into communities at multiple levels, each capturing a subset of interconnected entities. Finally, LLMs are used to generate detailed summaries for each community, creating a more organized and accessible representation of the underlying knowledge.

**Figure 2. The parital knowledge graph generated by LightRAG and Microsoft GraphRAG system.**

**Retrieval and answering.** In the retrieval phase, Microsoft GraphRAG develops different approaches to target different questions. For questions that rely on summarization and reasoning over multiple information sources, Microsoft GraphRAG applies a global search algorithm. The process begins by randomly shuffling all communities below a certain level (defined by the user). Next, the system sequentially combines the communities, divides them into different batches (one batch can contain multiple communities), and feeds each batch, along with the query, into the answering module. This module generates intermediate responses for each section and assigns a relevance score indicating how well the communities align with the query. Once all intermediate responses and their associated scores are collected, the system synthesizes a final answer by integrating the intermediate results. The answering module is usually implemented using the LLM API, ensuring robust natural language processing capabilities. For questions that focus on a few particular entities, the system adopts a local search algorithm. In local search, keywords will first be extracted from the query. Next, top entities, relationships, and communities that are relevant to the extracted keywords are retrieved and fed into the LLM as contextual knowledge. Finally, the LLM uses this context to directly answer the query.

### 3.3 LightRAG

LightRAG[7]. is a recently proposed GraphRAG framework designed to address several key limitations of traditional GraphRAG.

**Indexing.** LightRAG employs a similar approach to extract entities and relations from the original documents to construct a knowledge graph. However, it introduces a significant enhancement by enabling the incremental integration of new knowledge into an existing knowledge graph. Meanwhile, LightRAG will not generate communities for the query but will directly save the knowledge graph.

**Retrieval and answering.** In the query phase, the system processes a given question by first extracting keywords from it using LLMs. Particularly, the LightRAG design has three different retrieval modes: local, global, and hybrid. For **local mode**, the LLMs will extract low-level keywords from the query (like specific entities mentioned in the query). For the **global mode**, the LLMs will extract high-level keywords (like concepts related to the query). For **hybrid mode**, both low-level and high-level keywords will be included. After extraction, these keywords are then used to search the entire knowledge graph for matching entities and relations, along with their associated attributes or relationships. Subsequently, subgraphs centered around the identified entities and relations are extracted and passed to the answering module to generate a response.

### 3.4 Comparison between Microsoft GraphRAG, LightRAG, and AlzheimerRAG.

While both Microsoft GraphRAG and LightRAG are graph-based RAG systems, there are several key differences between the two approaches. First, in order to avoid performance bottlenecks on the retrieval phase, Microsoft GraphRAG organizes the knowledge graph into hierarchical clusters. This structure facilitates complex multi-hop reasoning but requires significant computational resources for graph construction and maintenance. Instead, LightRAG designed an innovative retrieval mechanism to efficiently retrieve knowledge directly from the knowledge graph. Further, since LightRAG only uses the raw knowledge graph, adding new information or removing incorrect data is straightforward. However, Microsoft GraphRAG requires re-indexing of the whole knowledge graph if you want to include new documents, making it costly to integrate new information to the model.

**(a) Comprehensiveness**

You are tasked with evaluating the comprehensiveness of two answers to a scientific question. Comprehensiveness is defined as how much detail the answer provides to cover all aspects and details of the question.

Here is the question: <question>{QUESTION}</question>

Here is the answer generated by the Graph RAG system:
<graph_rag_answer>
{GRAPH_RAG_ANSWER}
</graph_rag_answer>

Here is the answer generated by the Chat LLM model:
<chat_llm_answer>
{CHAT_LLM_ANSWER}
</chat_llm_answer>

Carefully analyze both answers, focusing on their comprehensiveness. Consider the following:
1. How thoroughly does each answer address all aspects of the question?
2. Which answer provides more relevant details and explanations?
3. Does either answer miss any important points related to the question?

Based on your analysis, choose the answer that demonstrates better comprehensiveness. Provide a brief explanation for your choice, highlighting the key factors that made the chosen answer more comprehensive.

Present your evaluation in the following format:
<evaluation>
<reasoning>
[Your explanation for why the chosen answer is more comprehensive]
</reasoning>
<choice>
[State which answer is more comprehensive: "Graph RAG" or "Chat LLM"]
</choice>
</evaluation>

**(b) Empowerment**

You are tasked with evaluating the potential of using a graph RAG system for scientific question answering compared to a normal chat model. Your role is to act as a judge and determine which of two given answers better demonstrates empowerment. Empowerment is defined as: How well does the answer help the reader understand and make informed judgments about the topic?

Here is the question that was asked: <question>{QUESTION}</question>

Here is the answer generated by the Graph RAG system: <graph_rag_answer>{GRAPH_RAG_ANSWER}</graph_rag_answer>

Here is the answer generated by the Chat LLM model: <chat_llm_answer>{CHAT_LLM_ANSWER}</chat_llm_answer>

To evaluate the empowerment of these two answers:

1. Carefully read and analyze both answers.
2. Consider how well each answer helps the reader understand the topic.
3. Assess how effectively each answer enables the reader to make informed judgments about the topic.
4. Compare the two answers based on their empowerment potential.

After your evaluation, provide your reasoning and final judgment. Your response should include:

1. A brief explanation of why you believe one answer demonstrates better empowerment than the other.
2. Your final judgment on which answer is better in terms of empowerment.

Present your evaluation in the following format:
<evaluation>
<reasoning>
[Your explanation for why the chosen answer is better]
</reasoning>
<choice>
[State which answer is more empowerment for user: "Graph RAG" or "Chat LLM"]
</choice>
</evaluation>

**Figure 3. Evaluation prompts used for evaluating comprehensiveness, empowerment, diversity, and directness.**

# 4 Evaluation of the performance of GraphRAG

## 4.1 Evaluation Protocol

In this section, we systematically evaluate the performance of GraphRAG for scientific questions. Specifically, we would like to answer the following questions. **Q1: Does GraphRAG provide better responses to scientific questions compared to standard LLMs? Q2: How do the different indexing and retrieval methods affect GraphRAGs performance?**

**Models:** We selected two representative GraphRAG methods, including Microsoft GraphRAG[8] and LightRAG[7]. For both methods, we employed GPT-4o[1] as the base LLM for both the indexing and answering modules to ensure a fair evaluation. For baseline, we included the GPT-4o API.

**Data:** To evaluate the performance of GraphRAG, we collected 50 papers focusing on Alzheimer's disease as the external database. The details of all collected papers can be found in **Supplement Table 1-2**. Next, we converted the pdf file of all papers into .txt documents and then constructed the database $\widehat{D}$ using two GraphRAG methods with the corresponding indexing function. In **Figure 2**, we visualized the generated knowledge graph from both Microsoft GraphRAG and LightRAG indexing modules.

**Evaluation procedure.** To conduct the evaluation, we first collected 70 questions related to Alzheimer's disease, curated from professional researchers in the field. The detailed question list can be found in **Supplement Table 3-4**. Next, we input all questions into both GraphRAG models and baseline GPT-4o API to generate answers. After we obtained answers, we evaluated the performance of both GraphRAG models by comparing it with the answer from the baseline GPT-4o API. Specifically, we evaluated the answer on the following aspects:

- **Comprehensiveness.** How much detail does the answer provide to cover all aspects and details of the question?

- **Diversity.** How varied and rich is the answer in providing different perspectives and insights on the question?

- **Empowerment.** How well does the answer help the researcher understand and make an informed judgment about the question?

- **Directness.** How specifically and clearly does the answer address the question?

Since the first four evaluation aspects do not assess the accuracy of the answers—an inherently challenging task without domain experts—we adopt the LLM-as-judge approach[19], similar to the evaluation methodology used in Microsoft GraphRAG and LightRAG. Specifically, we utilize the Claude 3.5 sonnet as the judge. For each answer pair (one generated by GraphRAG and the other by the baseline), Claude 3.5 is tasked with selecting the better answer based on the predefined criteria outlined in the prompt for each evaluation aspect. The detailed prompt used for these evaluations is provided in Figure 3.

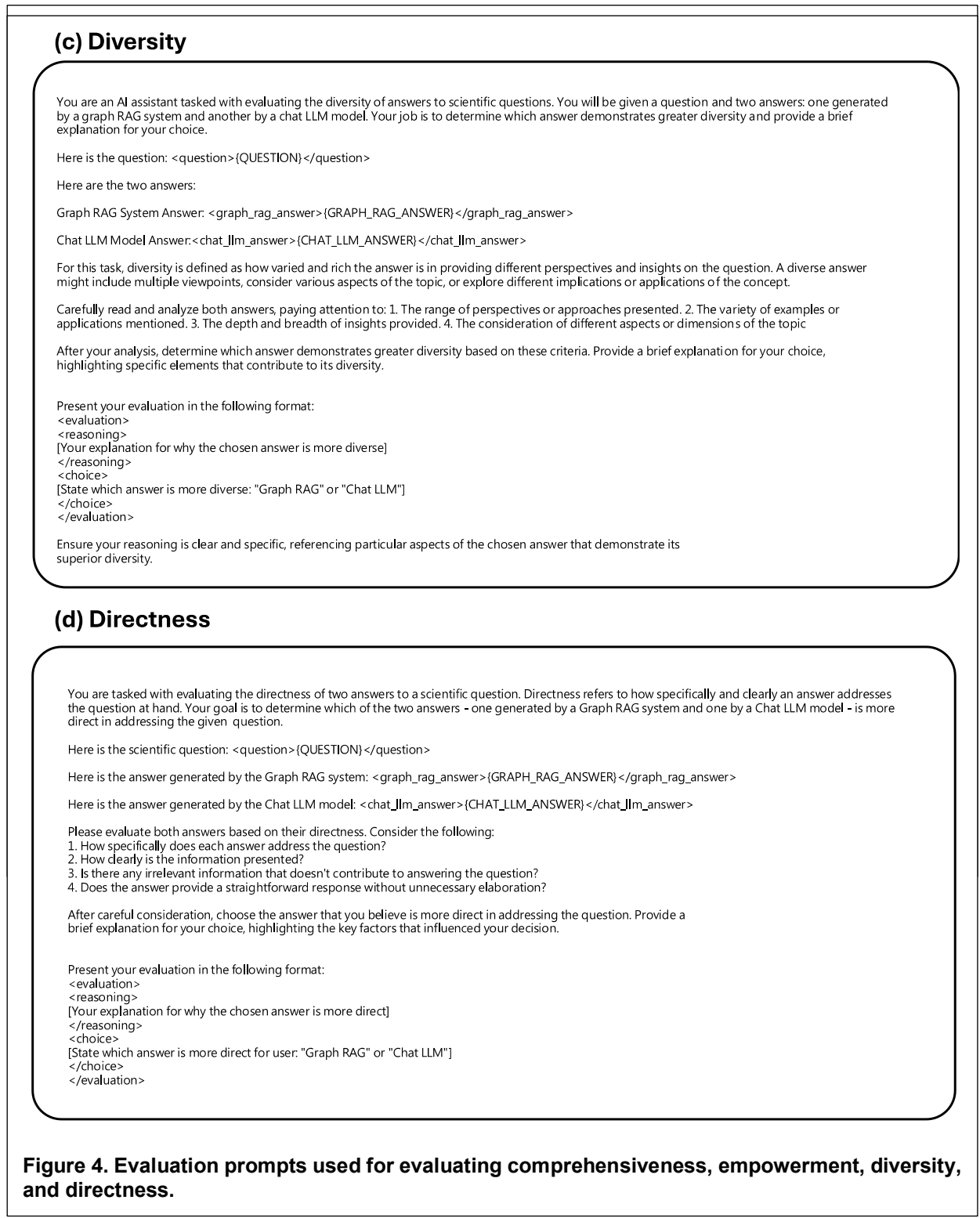

## (c) Diversity

You are an AI assistant tasked with evaluating the diversity of answers to scientific questions. You will be given a question and two answers: one generated by a graph RAG system and another by a chat LLM model. Your job is to determine which answer demonstrates greater diversity and provide a brief explanation for your choice.

Here is the question: <question>{QUESTION}</question>

Here are the two answers:

Graph RAG System Answer: <graph_rag_answer>{GRAPH_RAG_ANSWER}</graph_rag_answer>

Chat LLM Model Answer:<chat_llm_answer>{CHAT_LLM_ANSWER}</chat_llm_answer>

For this task, diversity is defined as how varied and rich the answer is in providing different perspectives and insights on the question. A diverse answer might include multiple viewpoints, consider various aspects of the topic, or explore different implications or applications of the concept.

Carefully read and analyze both answers, paying attention to: 1. The range of perspectives or approaches presented. 2. The variety of examples or applications mentioned. 3. The depth and breadth of insights provided. 4. The consideration of different aspects or dimensions of the topic

After your analysis, determine which answer demonstrates greater diversity based on these criteria. Provide a brief explanation for your choice, highlighting specific elements that contribute to its diversity.

Present your evaluation in the following format:
<evaluation>
<reasoning>
[Your explanation for why the chosen answer is more diverse]
</reasoning>
<choice>
[State which answer is more diverse: "Graph RAG" or "Chat LLM"]
</choice>
</evaluation>

Ensure your reasoning is clear and specific, referencing particular aspects of the chosen answer that demonstrate its superior diversity.

## (d) Directness

You are tasked with evaluating the directness of two answers to a scientific question. Directness refers to how specifically and clearly an answer addresses the question at hand. Your goal is to determine which of the two answers - one generated by a Graph RAG system and one by a Chat LLM model - is more direct in addressing the given question.

Here is the scientific question: <question>{QUESTION}</question>

Here is the answer generated by the Graph RAG system: <graph_rag_answer>{GRAPH_RAG_ANSWER}</graph_rag_answer>

Here is the answer generated by the Chat LLM model: <chat_llm_answer>{CHAT_LLM_ANSWER}</chat_llm_answer>

Please evaluate both answers based on their directness. Consider the following:
1. How specifically does each answer address the question?
2. How clearly is the information presented?
3. Is there any irrelevant information that doesn't contribute to answering the question?
4. Does the answer provide a straightforward response without unnecessary elaboration?

After careful consideration, choose the answer that you believe is more direct in addressing the question. Provide a brief explanation for your choice, highlighting the key factors that influenced your decision.

Present your evaluation in the following format:
<evaluation>
<reasoning>
[Your explanation for why the chosen answer is more direct]
</reasoning>
<choice>
[State which answer is more direct for user: "Graph RAG" or "Chat LLM"]
</choice>
</evaluation>

**Figure 4. Evaluation prompts used for evaluating comprehensiveness, empowerment, diversity, and directness.**

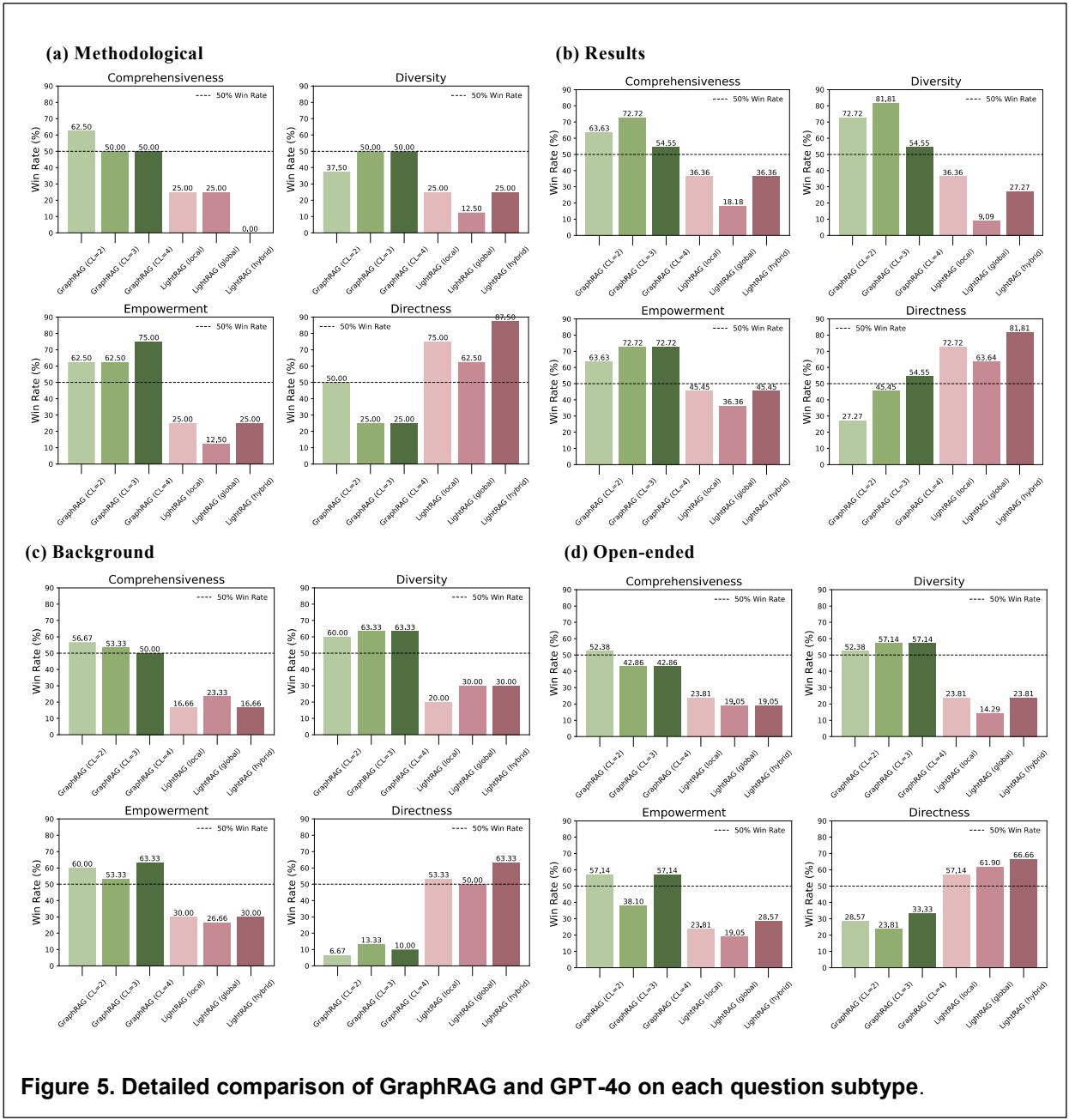

**Figure 5. Detailed comparison of GraphRAG and GPT-4o on each question subtype**.

## 4.2 Evaluation results

The evaluation results of the first four aspects are presented in **Figure 3**. In **Table 1** and **Table 2**, we also provide example answers for both Microsoft GraphRAG and LightRAG, compared to baseline GPT-4o. Specifically, for Microsoft GraphRAG, we only evaluate the global search mode, as most of our designed questions require summarization from multiple knowledge sources. To ensure the robustness of the evaluation, we include the comparison using different community levels (denoted CL). Given a certain community level $c$, the GraphRAG system will retrieve all community reports equal or below $c$.. Since smaller community levels tend to produce more abstract summaries, while larger levels incorporate more detailed information, it is expected that the results will vary depending on the community level. After indexing of 50 papers, there are 67, 298, 444, 147, and 10 different communities from level 0 to level 4 respectively. The communities at a larger level are more specific and describe more detailed knowledge. For LightRAG, we simultaneously compared the answers generated from all three different search modes.

From our results, we derive the following key conclusions. First, GraphRAG consistently generates more comprehensive, diverse, and empowering responses compared to standard LLM chatbots, as evidenced by its win rate exceeding 50%. This finding supports the hypothesis that incorporating contextual knowledge relevant to a query enhances the quality of LLM-generated responses. However, we observe no major difference in performance across different community levels.

Second, LightRAG demonstrates superior performance over standard LLMs only in terms of directness, while underperforming in comprehensiveness, diversity, and empowerment in most cases. A likely explanation is that the effectiveness of LightRAG is highly dependent on the accuracy of its retrieval step, specifically the keyword extraction process. Since LightRAG relies solely on an LLM for keyword selection, it may fail to identify the most informative terms. For example, in many cases, LightRAG extracts Alzheimer's disease as a keyword—while relevant, this term does not provide sufficient granularity to pinpoint the most useful information within the database, as all stored knowledge pertains to Alzheimer's disease. In contrast, Microsoft GraphRAG avoids keyword extraction altogether. Instead, it iterates over all communities in the knowledge base, ensuring access to the most pertinent information. However, this approach presents a scalability challenge: as the database grows (e.g., to tens of thousands of papers), the increasing number of communities would significantly amplify computational costs due to exhaustive iteration. Using our experiment as an example, all 50 papers created 67, 298, and 444 community reports on levels 0, 1, and 2, respectively. Suppose each community has, on average, 500 tokens for the description. GPT-4o costs 2.5 dollars for 1 million input tokens, which means that using reports at and below community level 2 to answer a single question will cost about 0.9 dollar extra compared to using community level 0 only on generating intermediate answers. These estimates also do not count the extra cost induced in outputting intermediate answers, and summarizing over these intermediate answers, as the number of intermediate answers to generate is also increased due to the increased number of communities.

Finally, despite its simplicity and cost efficiency, the ordinary LLM chatbot remains competitive with the GraphRAG systems especially LightRAG, as shown in **Figure 4**. Several factors may contribute to this outcome. First, modern LLMs are continuously updated, incorporating an expanding body of knowledge through both pre-training and post-training processes. As a result, many questions can be answered effectively without the need for additional external knowledge. Second, RAG-based systems primarily prompt the LLM to generate responses based on retrieved contextual information. While this approach ensures grounding in relevant documents, it may also constrain the model's ability to leverage its built-in knowledge. By focusing exclusively on extracting useful insights from the provided context, RAG-based systems may inadvertently limit the model's flexibility in synthesizing broader knowledge from its training corpus, especially when the database in the RAG-based system is incomplete.

To further evaluate how GraphRAG performs on different types of questions, we divided all questions into four subtypes: Methodological, Results, Background, and Open-ended. In Figure 5, we show the comparison results of GraphRAG to GPT-4o on each different question subtype. There are several interesting observations: First, both Microsoft GraphRAG and LightRAG obtain significantly better performance on subtype Results, compared to their performance on other subtypes. The reason could be that these questions usually ask about specific results that exist in a particular paper, which may not be included or effectively memorized by LLM during the training. Instead, for GraphRAG, since we explicitly include these results in the graph database, the system can answer such questions much better than standard LLMs. Second, Microsoft GraphRAG consistently achieves better performance than GPT-4o on empowerment across different question subtypes and different community levels. Recall that the empowerment is how well the answer helps the researcher understand and make an informed judgment about the question. This result indicates that the use of community summarization may be used as an additional reasoning step for LLMs to digest different information and summarize it into knowledge that can be better leveraged by humans. Third, LightRAG achieved worse results comthan GPT-4o for each subtype on all metrics (except directness), which indicates that a better retrieval and prompting design is required for LightRAG to have better performance.

## 5 Analyzing the traceability of RAG and GraphRAG

In this section, we investigate the traceability of different RAG and GraphRAG systems. Traceability refers to the ability to track and verify the provenance of the information used to generate a response. A highly traceable answer should explicitly link its claims to specific sources, references, or intermediate reasoning steps, enabling users to assess its reliability, accuracy, and potential biases. In scientific research, high traceability is particularly desirable; for instance, a system that can accurately cite the specific paper and paragraph supporting its response would be highly valued. However, most ordinary LLMs struggle with traceability, often generating fabricated or inaccurate references due to hallucinations. In contrast, RAG-based systems inherently improve traceability by grounding their responses in external knowledge sources. Despite this advantage, there has been no systematic evaluation of the traceability of existing RAG and GraphRAG systems. In this paper, we take the first step in addressing this gap by evaluating traceability from both qualitative and quantitative perspectives in terms of scientific research.

### 5.1 Qualitative analysis of traceability

First, RAG-based systems adopt diverse design approaches. Some systems keep incorporating traceability when designing their method, while others do not. This variation makes direct comparison of traceability across systems challenging. To address this, we begin with a qualitative analysis by defining different levels of traceability in the context of scientific research. We then classify the traceability of various systems based on a key criterion: the extent to which a system can provide reliable traceability without significant modifications to its original method. For instance, some RAG systems were not designed with traceability in mind. However, if minor adjustments—such as modifying data input or reformatting prompts—enable them to generate accurate references, we still consider them traceable. We begin the discussion by first dividing traceability into four different levels: non-traceable, cluster level, multi-paragraph level, and single-paragraph level.

**Non-traceable:** A system that cannot provide any references for its answers is classified as non-traceable. Typically, machine learning and deep learning models, including LLMs, fall into this category. Here, we refer specifically to standard LLMs, such as the GPT API version. However, when accessed through a web interface, LLMs may include special add-ons such as web browsing, that enable them to retrieve and process external information, thereby introducing the possibility of traceability. Since closed-source models do not disclose how these add-ons are integrated on the backend, we do not analyze them in detail.

**Cluster-level:** Cluster-level traceability refers to a system's ability to provide references to a cluster of papers or documents rather than specific sources. A typical example of this approach is the global search mode of Microsoft GraphRAG. In this mode, the system iterates over all community reports to identify the most relevant reports for answering a query. However, each community report is generated by summarizing information from a cluster of nodes and relations, which themselves originate from multiple sources. As a result, it becomes nearly impossible to trace an answer back to a specific paper or paragraph. In general, any system that processes raw knowledge through summarization, pruning, or other refinement steps before generating responses falls into this category.

**Multi-paragraph level:** A more granular level of traceability than the cluster level is the multi-paragraph level. Systems in this category can reference specific documents and paragraphs rather than entire clusters of papers. However, since each referenced source may be synthesized from multiple paragraphs in the original documents, it remains difficult to pinpoint the exact paragraph used to generate an answer. Most GraphRAG-based systems, such as the local search mode in Microsoft GraphRAG or LightRAG, fall into this category. These methods first extract relevant nodes and relations from a constructed knowledge graph and use this refined information to generate responses. As a result, they can readily output referenced entities. However, because each entity may be extracted independently from multiple sources, tracing the most relevant source for a given query remains challenging. For example, the keyword "APOE" may be extracted from multiple papers on Alzheimer's disease, making it difficult to determine which source is the most pertinent to a specific question. However, for keywords that are more detailed or only exist in a single source, these systems can provide more accurate referencing.

**Single-paragraph level:** The most fine-grained level of traceability is the single-paragraph level, where a system can directly reference a specific paragraph in the raw knowledge base as the source of its response. Systems that achieve this level provide the highest degree of transparency, making it easier to verify claims and assess the reliability of generated answers. Standard RAG systems like PaperQA Lála et al. [2023], Skarlinski et al. [2024], with slight modifications, such as adjusting retrieval mechanisms or refining prompt formatting, can typically provide references at this level. In some cases, GraphRAG-based systems can also achieve single-paragraph traceability, particularly when an entity or concept appears in only one source. However, as knowledge graphs often aggregate and abstract information from multiple documents, direct one-to-one mappings between a response and an individual paragraph are rare. This challenge becomes even more pronounced as the size of the database grows, increasing the likelihood that information is spread across multiple sources. On the other hand, some questions are innately so broad that a single paragraph-level answer is not possible. As the systems continue to grow and gain more knowledge for a particular domain, like Alzheimer's disease, more references about the same aspect will be used to answer the question.

**Additional discussion on the trade-off between traceability and performance:** In the context of scientific research, traceability typically refers to a system's ability to trace an answer back to the exact raw source, such as a specific paragraph in a paper. However, modern RAG-based systems often prioritize improving performance and accuracy over strict traceability. For instance, GraphRAG enhances retrieval quality by transforming raw documents into structured graphs, effectively filtering out redundant, noisy, or incorrect information while preserving the most valuable insights. This approach not only refines the knowledge representation but also improves retrieval performance compared to standard vector-based methods. However, this refinement comes with a trade-off. Higher processing of raw data often reduces traceability. Systems that impose strict retrieval constraints, such as retrieving only the most relevant paragraph instead of multiple related passages, tend to achieve finer-grained traceability. Yet, this often results in the loss of valuable contextual information that could enhance response accuracy and completeness. Balancing both high accuracy and high traceability remains an open challenge. While enhanced knowledge structuring improves reliability and retrieval effectiveness, it also introduces complexities in tracing answers back to their precise origins.

## 6   A simple interface for using RAG with the Alzheimer's disease knowledge base

To further facilitate the use of RAG and GraphRAG in the field of Alzheimer's disease, we constructed a simple interface for using RAG and GraphRAG. The interface is built on the open-source repository kotaemon[20]. The interface includes standard RAG, Microsoft GraphRAG, and LightRAG systems. An overview of the interface is shown in **Figure 6**.

The interface contains three modules. On the left, the user can review the conversation history. Meanwhile, the user can select the method they want to use for answering the question. The File collection, GraphRAG collection, and LightRAG collection mean standard RAG, Microsoft GraphRAG, and LightRAG, respectively. The interface also provides a protocol for directly uploading new documents. In the middle, the user can view the current conversation and ask new questions. The interface also provides options for including references and citations for the answer. This part is implemented by the kotaemon. Specifically, given the answer and context provided to the LLM, the interface will call the LLM again and ask which part of the context has been referenced for the answer. On the right, the reference will be provided if be included. Given the different method, different references will be provided as the context are different. For standard RAG, the raw paper section will be shown. For Microsoft GraphRAG, the relevant communities are provided. For LightRAG, the related entities and relations are provided.

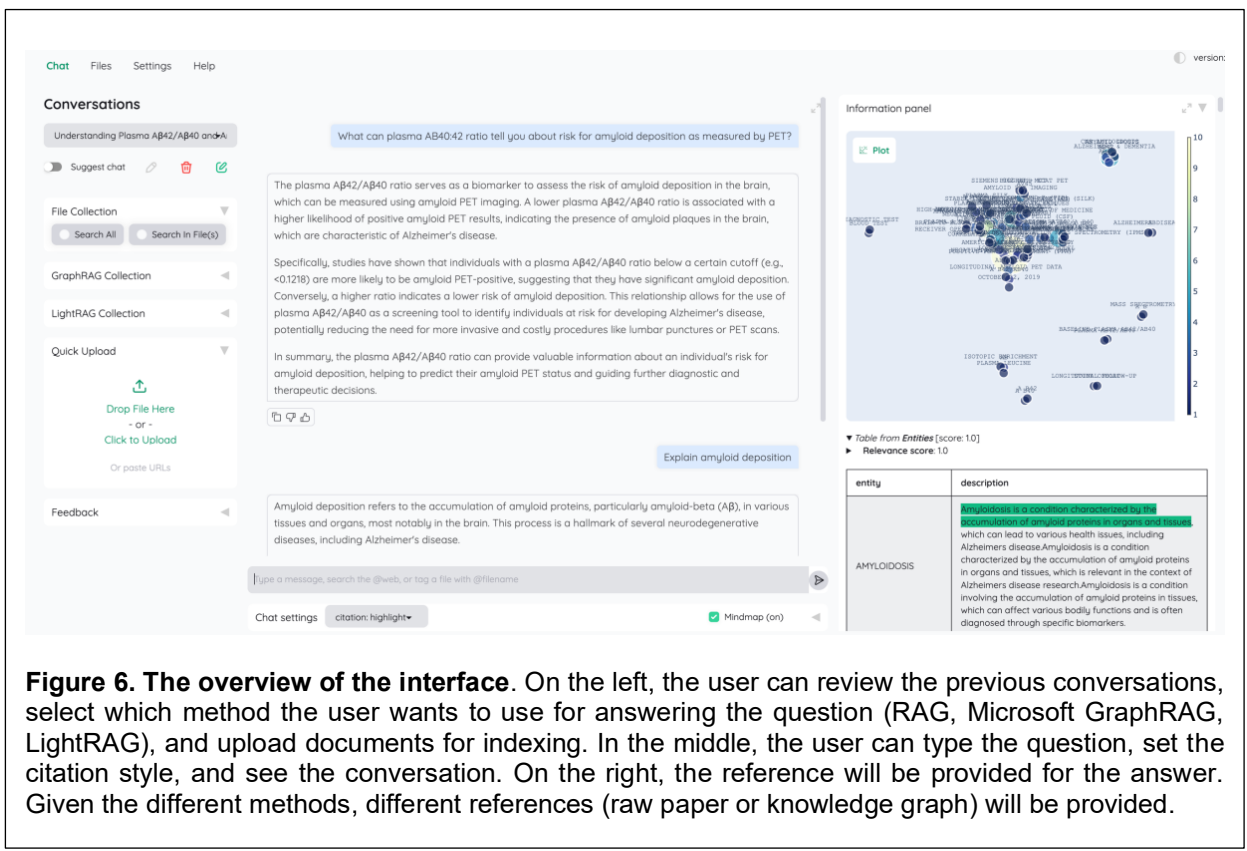

**Figure 6. The overview of the interface**. On the left, the user can review the previous conversations, select which method the user wants to use for answering the question (RAG, Microsoft GraphRAG, LightRAG), and upload documents for indexing. In the middle, the user can type the question, set the citation style, and see the conversation. On the right, the reference will be provided for the answer. Given the different methods, different references (raw paper or knowledge graph) will be provided.

Based on this interface, we preprocessed 50 papers related to Alzheimer's disease (the same as the evaluation paper we used for all three methods so that users can directly ask questions related to Alzheimer's disease.

## 7 Conclusions

In this study, we conducted a systematic evaluation of the potential of the GraphRAG system for Alzheimer's disease. Specifically, we collected 50 papers in the field of Alzheimer's disease and constructed the graph database using two representative GraphRAG systems, namely Microsoft GraphRAG and LightRAG. We also collect 70 questions from professional researchers in the field to assess the ability of the GraphRAG system. The results demonstrate the potential of GraphRAG to enhance the answer of LLMs for scientific questions. We also analyze the current traceability of different RAG system and provide a. holistic view. Finally, we build up a simple interface for researchers in the alzheimer's disease field to test different RAG system.

**Table 1: Example question with generated answers from Microsoft Graph RAG (CL=2) and baseline, as well as LLM-generated assessments**.

| Question | What amyloid beta species have been measured by stable isotope labeled kinetics (SILK)? |
| --- | --- |

**Graph RAG**

**Overview of SILK in Amyloid Beta Research**

Stable Isotope Labeling Kinetics (SILK) is a sophisticated method used to study the dynamics of amyloid-beta (A$\beta$) peptides, which are crucial in Alzheimer's disease research. This technique involves isotopic labeling to analyze the production, clearance, and turnover rates of different A$\beta$ species, providing valuable insights into their metabolic processes.

**Measured Amyloid Beta Species**

The SILK method has been employed to measure several amyloid-beta isoforms, specifically A$\beta$38, A$\beta$40, and A$\beta$42. These measurements are conducted in both cerebrospinal fluid (CSF) and plasma, offering a comprehensive view of amyloid-beta metabolism in the human body.

**Significance in Alzheimer's Disease**

The ability to measure these specific A$\beta$ species is significant for understanding the pathophysiology of Alzheimer's disease. By quantifying the kinetics of these peptides, researchers can gain insights into their production and clearance rates, which are critical for developing therapeutic strategies and understanding disease progression.

**Conclusion**

In summary, the SILK method is a powerful tool in Alzheimer's research, enabling the detailed study of amyloid-beta species such as A$\beta$38, A$\beta$40, and A$\beta$42. These insights are essential for advancing our understanding of neurodegenerative diseases and developing effective treatments.

**GPT-4o**

Stable Isotope Labeling Kinetics (SILK) is a powerful technique used for measuring the synthesis and clearance rates of various amyloid-beta (A$\beta$) species. This methodology has been applied to study the dynamics of A$\beta$ in both cerebrospinal fluid (CSF) and blood plasma, providing valuable insights into Alzheimer's disease pathology and progression.

**1. A$\beta$40 and A$\beta$42**

These are the most commonly studied species in SILK experiments. A$\beta$40 and A$\beta$42 differ by two amino acids, with A$\beta$42 being more prone to aggregation and more closely associated with neurotoxic effects in Alzheimer's disease. Studies utilizing SILK have measured the production and clearance rates of these peptides, providing critical information about their kinetics in vivo.

**Total A$\beta$ (including multiple isoforms)**

Some studies aim to assess the overall turnover of A$\beta$ species collectively, without distinguishing between individual isoforms. While specific measurements may target the most prevalent species like A$\beta$40 and A$\beta$42, analyzing total A$\beta$ kinetics can help in understanding the general behavior of A$\beta$-related processes in the body.

These measurements are crucial in Alzheimer's research because alterations in the kinetics of A$\beta$, particularly decreased clearance or increased production of A$\beta$42, are believed to contribute to the pathological aggregation seen in the disease. Through SILK, researchers can gain a clearer picture of how these kinetic changes relate to disease states, potentially guiding the development of therapeutic interventions.

**LLM Decision**

**Comprehensiveness: Winner=GraphRAG**

The Graph RAG answer explicitly identifies three specific A$\beta$ species (A$\beta$38, A$\beta$40, and A$\beta$42) that have been measured using SILK, while the Chat LLM answer only mentions A$\beta$40 and A$\beta$42, missing A$\beta$38.

**Diversity: Winner=GraphRAG**

The Graph RAG answer demonstrates greater diversity for several key reasons. 1. Structural Organization: It presents information through multiple distinct sections (Overview, Measured Species, Significance, Conclusion), offering varied perspectives on the topic. 2. Broader Species Coverage: It specifically mentions three different A$\beta$ species (A$\beta$38, A$\beta$40, and A$\beta$42), while the Chat LLM answer primarily focuses on A$\beta$40 and A$\beta$42. 3. Contextual Integration: It discusses both measurement environments (CSF and plasma) while connecting this to broader research implications and therapeutic development.

**Empowerment: Winner=GraphRAG**

The Graph RAG answer better empowers readers by providing more comprehensive, well-organized information with verifiable sources, enabling them to make more informed judgments about the topic and potentially pursue further research.

**Directness: Winner=GraphRAG**

The Chat LLM answer is less direct because: 1. It only mentions A$\beta$40 and A$\beta$42, missing A$\beta$38. 2. It includes speculative discussion about aggregation and neurotoxic effects that wasn't asked for. The information about specific species is embedded within longer explanatory paragraphs rather than presented directly. While both answers provide context about SILK methodology and its importance in Alzheimer's research, the Graph RAG answer delivers the specific requested information more clearly and directly, with less peripheral discussion.

**Table 2: Example question with generated answers from LightRAG (hybrid mode) and baseline, as well as LLM-generated assessments.**

| Question | Describe the various isoforms of APOE. |
|---|---|

**LightRAG**

**APOE Isoforms Overview**: Apolipoprotein E (APOE) has three primary isoforms: ApoE2, ApoE3, and ApoE4. These isoforms differ slightly by their amino acid sequences but have significant implications for lipid metabolism and neurodegenerative diseases, particularly Alzheimer's disease.

**ApoE2**: ApoE2 is characterized by the cysteine residues at positions 112 and 158. It is the least common allele and is considered to confer some protective effects against Alzheimer's disease due to its better binding affinity for amyloid-beta, facilitating greater clearance compared to ApoE4. ApoE2 has a lower risk association with Alzheimer's disease compared to the other isoforms.

**ApoE3**: The most prevalent isoform, ApoE3, has a cysteine at position 112 and an arginine at position 158. It is generally associated with normal risk levels for Alzheimer's and is considered neutral concerning the disease. ApoE3 is the standard isoform used for experimental comparisons and is noted for efficiently binding and clearing amyloid-beta in the brain.

**ApoE4**: ApoE4 is associated with a higher risk of Alzheimer's disease. It includes arginine at both positions 112 and 158. The structural differences in ApoE4 lead to less effective amyloid-beta clearance, contributing to amyloid plaque formation. This isoform significantly impacts disease risk and progression, influencing amyloid deposition and tau pathology.

**Functional Implications**: The different isoforms of apoE affect the concentration and clearance of amyloid-beta, thereby influencing Alzheimer's disease pathology. ApoE4, in particular, is linked to impaired clearance and increased brain amyloid deposition, acting as a major genetic risk factor for Alzheimer's disease. Meanwhile, ApoE2 is linked to potential protective effects, whereas ApoE3 maintains a neutral profile regarding Alzheimer's risk.

**GPT-4o**

Apolipoprotein E (APOE) is a class of proteins involved in the metabolism of fats in the body and is crucial in the transport and clearance of lipoproteins from the bloodstream. APOE exists in several isoforms that differ due to genetic variations in the APOE gene.

**ApoE2**: Genetic Variant: It arises from a change in the APOE gene where cysteine is present at both residues 112 and 158 (Cys112/Cys158). Lipid Metabolism: APOE2 is associated with lower levels of total cholesterol and LDL cholesterol. It can result in a condition known as Type III hyperlipoproteinemia when found in individuals who are homozygous for the APOE2 allele. Neurological Impact: This allele is relatively rare compared to the other two and is associated with a decreased risk of Alzheimer's disease.

**ApoE3**: Genetic Variant: The most common isoform, APOE3, has cysteine at position 112 and arginine at position 158 (Cys112/Arg158). Lipid Metabolism: Considered the neutral isoform, it functions normally in lipid transport and does not significantly influence cholesterol levels or risk for Alzheimer's compared to APOE2 and APOE4. Neurological Impact: It is regarded as the standard or baseline against which the other isoforms are compared, particularly in medical research.

**ApoE4**: Genetic Variant: APOE4 results from having arginine at both residues 112 and 158 (Arg112/Arg158). Lipid Metabolism: This isoform is linked to higher levels of LDL cholesterol and total cholesterol, contributing to cardiovascular diseases. Neurological Impact: APOE4 is a known genetic risk factor for Alzheimer's disease. Individuals carrying one APOE4 allele have a higher risk, and those with two copies have an even greater risk compared to carriers of APOE3 or APOE2 alleles.

**Significance in Research and Medicine**: Understanding the variations and their implications not only help in genetic counseling and risk assessment but also potentially pave the way for the development of targeted treatments for neurodegenerative and cardiovascular diseases.

**LLM Decision**

**Comprehensiveness: Winner=Chat LLM**
The Graph RAG answer focuses primarily on this aspect and lacks detail about other important functions and implications of these proteins, particularly their role in lipid metabolism.

**Diversity: Winner=Chat LLM**
While the Graph RAG answer is detailed, it focuses primarily on Alzheimer's disease and amyloid-beta interactions, missing the broader implications and diverse clinical manifestations that the Chat LLM answer captures.

**Empowerment: Winner=LightRAG**
The Chat LLM answer provides more general information and fewer specific details that would enable readers to make informed judgments about the molecular and functional differences between APOE isoforms.

**Directness: Winner=LightRAG**
The Graph RAG answer stays more tightly focused on describing the isoforms themselves. Its organization makes it easier to understand the distinct characteristics of each isoform, which is the primary focus of the question.

**Supplement Table 1: The detailed 50 papers included in the external database**.

| Index | Question |
|---|---|
| 1 | Amyloid $\beta$ concentrations and stable isotope labeling kinetics of human plasma specific to central nervous system amyloidosis |
| 2 | High-precision plasma $\beta$-amyloid 42/40 predicts current and future brain amyloidosis |
| 3 | Apolipoprotein E O-glycosylation is associated with amyloid plaques and APOE genotype |
| 4 | Baseline levels and longitudinal changes in plasma A$\beta$42/40 among Black and white individuals |
| 5 | Predicting continuous amyloid PET values with CSF and plasma A$\beta$42/A$\beta$40 |
| 6 | Decreased Clearance of CNS $\beta$-Amyloid in Alzheimer's Disease |
| 7 | Increased in Vivo Amyloid-$\beta$42 Production, Exchange, and Loss in Presenilin Mutation Carriers |
| 8 | Age and amyloid effects on human central nervous system amyloid-beta kinetics |
| 9 | [DT-01–03]: Concentrations and stable isotope label kinetics of human plasma amyloid beta. |
| 10 | Tau Phosphorylation Rates Measured by Mass Spectrometry Differ in the Intracellular Brain vs. Extracellular Cerebrospinal Fluid Compartments and Are Differentially Affected by Alzheimer's Disease |
| 11 | Longitudinal associations between exercise and biomarkers in autosomal dominant Alzheimer's disease |
| 12 | Investigation of sex differences in mutation carriers of the Dominantly Inherited Alzheimer Network |
| 13 | Dominantly inherited Alzheimer's disease in Latin America: Genetic heterogeneity and clinical phenotypes |
| 14 | CAP—advancing the evaluation of preclinical Alzheimer disease treatments |
| 15 | Suvorexant Acutely Decreases Tau Phosphorylation and A$\beta$ in the Human CNS |
| 16 | A soluble phosphorylated tau signature links tau, amyloid and the evolution of stages of dominantly inherited Alzheimer's disease |
| 17 | Alzheimer's disease |
| 18 | Amyloid-$\beta$ Dynamics Are Regulated by Orexin and the Sleep-Wake Cycle |
| 19 | Head-to-head comparison of 10 plasma phospho-tau assays in prodromal Alzheimer's disease |
| 20 | Human apoE Isoforms Differentially Regulate Brain Amyloid-$\beta$ Peptide Clearance |
| 21 | SILK studies - capturing the turnover of proteins linked to neurodegenerative diseases |
| 22 | Astrocytic 4R tau expression drives astrocyte reactivity and dysfunction |
| 23 | Recapitulation of endogenous 4R tau expression and formation of insoluble tau in directly reprogrammed human neurons |
| 24 | Modeling late-onset Alzheimer's disease neuropathology via direct neuronal reprogramming |
| 25 | Clinical and Biomarker Changes in Dominantly Inherited Alzheimer's Disease |

| Index | Question |
|---|---|
| 26 | Tau Kinetics in Neurons and the Human Central Nervous System |
| 27 | CSF tau microtubule binding region identifies tau tangle and clinical stages of Alzheimer's disease |
| 28 | CSF tau microtubule-binding region identifies pathological changes in primary tauopathies |
| 29 | Spatial patterns of neuroimaging biomarker change in individuals from families with autosomal dominant Alzheimer's disease: a longitudinal study |
| 30 | Tau PET in autosomal dominant Alzheimer's disease: relationship with cognition, dementia and other biomarkers |
| 31 | Human amyloid-$\beta$ synthesis and clearance rates as measured in cerebrospinal fluid in vivo |
| 32 | Sleep-wake cycle and diurnal fluctuation of amyloid-$\beta$ as biomarkers of brain amyloid pathology |
| 33 | Longitudinal change in CSF biomarkers in autosomal-dominant Alzheimer's disease |
| 34 | Blood plasma phosphorylated-tau isoforms track CNS change in Alzheimer's disease |
| 35 | Effect of sleep on overnight cerebrospinal fluid amyloid $\beta$ kinetics |
| 36 | Effects of age and amyloid deposition on A$\beta$ dynamics in the human central nervous system |
| 37 | Amyloid-$\beta$ efflux from the central nervous system into the plasma |
| 38 | Diversity of Amyloid-beta Proteoforms in the Alzheimer's Disease Brain |
| 39 | Statistical considerations when estimating time-saving treatment effects in Alzheimer's disease clinical trials |
| 40 | Application of machine learning to predict amyloid, metabolic, and structural neuroimaging biomarkers in the progression of autosomal dominant Alzheimer disease |
| 41 | Predicting sporadic Alzheimer's disease progression via inherited Alzheimer's disease-informed machine-learning |
| 42 | CSF tau phosphorylation occupancies at T217 and T205 represent improved biomarkers of amyloid and tau pathology in Alzheimer's disease |
| 43 | Highly accurate blood test for Alzheimer's disease is similar or superior to clinical cerebrospinal fluid tests |
| 44 | MAPT R406W increases tau T217 phosphorylation in absence of amyloid pathology |
| 45 | The performance of plasma amyloid beta measurements in identifying amyloid plaques in Alzheimer's disease: a literature review |
| 46 | CSF proteomics identifies early changes in autosomal dominant Alzheimer's disease |
| 47 | Fluid biomarkers in the context of amyloid-targeting disease-modifying treatments in Alzheimer's disease |
| 48 | A genetic and proteomic comparison of key AD biomarkers across tissues |
| 49 | Plasma Phosphorylated Tau 217 and A$\beta$42/40 to Predict Early Brain A$\beta$ Accumulation in People Without Cognitive Impairment |
| 50 | Downstream Biomarker Effects of Gantenerumab or Solanezumab in Dominantly Inherited Alzheimer DiseaseThe DIAN-TU-001 Randomized Clinical Trial |

**Supplement Table 3: The used 70 questions for evaluation.**

| Index | Question |
|---|---|
| 1 | What amyloid beta species have been measured by stable isotope labeled kinetics (SILK)? |
| 2 | Describe the turnover kinetics of various amyloid beta species relative to one another. |
| 3 | How do the kinetics of amyloid beta species change in people with amyloid deposition? |
| 4 | What can plasma A$\beta$B40:42 ratio tell you about risk for amyloid deposition as measured by PET? |
| 5 | Describe the various isoforms of APOE. |
| 6 | How do amino acid changes in the isoforms of APOE affect its function? |
| 7 | Please define late onset Alzheimer's disease. |
| 8 | How does amyloid beta production and clearance affect late onset Alzheimer's disease progression and risk? |
| 9 | What is are the biggest risk factors for late onset Alzheimer's disease? |
| 10 | How do Alzheimer's disease risk factors affect the kinetics of amyloid beta isoforms? |
| 11 | What is known about tau and phosphorylation of tau in the context of Alzheimer's disease? |
| 12 | What is the mechanism of abnormal tau phosphorylation in Alzheimer's disease? |
| 13 | Does exercise affect Alzheimer's disease? |
| 14 | What biomarkers can be used to measure the effect of exercise on Alzheimer's disease? |
| 15 | What are the mechanisms by which exercise may slow or accelerate Alzheimer's disease? |
| 16 | How does amyloid beta affect tauopathy? |
| 17 | What is known about the mechanisms of synaptic transfer of tau? |
| 18 | What are the stages of Alzheimer's disease progression? |
| 19 | What biomarkers and tests can be used to identify Alzheimer's disease? |
| 20 | In what order are Alzheimer's biomarkers detectable leading up to the age of onset? |
| 21 | What is the effect of sleep on amyloid beta levels in the brain? |
| 22 | Do specific isoforms of amyloid beta differentially change with sleep deprivation? |
| 23 | What is the role of tau in the brain? |
| 24 | What gene expresses tau? |
| 25 | Describe the different forms of tau and how the relate to Alzheimer's disease. |
| 26 | How do the kinetics of tau isoforms change in Alzheimer's disease? |
| 27 | What relationship does tau have with amyloid plaque burden? |
| 28 | How do tau aggregates change with Alzheimer's disease progression? |
| 29 | What are tau aggregates composed of? |
| 30 | What are the differences in biomarkers between sporadic Alzheimer's and dominantly-inherited Alzheimer's? |
| 31 | What animal models are used to model Alzheimer's disease? |
| 32 | What are the strengths and weaknesses of various models of Alzheimer's disease? |
| 33 | What factors confound mass spectrometry based tests? |
| 34 | What causes amyloid deposition in tissue? |
| 35 | How does amyloid pathology spread throughout the brain? |

**Supplement Table 4: The used 70 questions for evaluation (Continued).**

| Index | Question |
|-------|----------|
| 36 | What proteins interact with amyloid plaques? |
| 37 | What proteins interact with neurofibrillary tangles? |
| 38 | What are the mechanisms for anti-amyloid treatments? |
| 39 | What are the best performing biomarker tests for amyloidosis in plasma and CSF? |
| 40 | Why are kinetics useful in studying Alzheimer's disease? |
| 41 | What age-related mechanisms lead to amyloid accumulation? |
| 42 | What are the advantages of measuring proteins by mass spectrometry? |
| 43 | Why does the ratio of amyloid beta 40:42 change in Alzheimer's disease? |
| 44 | How do mutations in tau affect it's aggregation? |
| 45 | What are the side effects of anti-amyloid treatments? |
| 46 | Do anti-amyloid treatments alter disease course or cognitive function? |
| 47 | How can artificial intelligence be leveraged to advance our understanding of Alzheimer's disease? |
| 48 | What protein modifications of amyloid-beta and tau are associated with Alzheimer's disease? |
| 49 | What is the function of amyloid-beta? |
| 50 | What are the major limitations of increasing access to blood tests for Alzheimer's disease? |
| 51 | Which animals get Alzheimer's disease? |
| 52 | Why don't old sharks get amyloid deposition? |
| 53 | What environmental factors increase the risk for developing Alzheimer's disease? |
| 54 | How does Alzheimer's pathology increase oxidative stress? |
| 55 | How do astrocytes affect Alzheimer's pathology? |
| 56 | What is the role of microglia in Alzheimer's pathology? |
| 57 | What is the minimum brain plaque load that can be measured by amyloid PET? |
| 58 | Are average CSF and plasma amyloid beta levels different across the world? |
| 59 | Does socioeconomic status alter the risk for Alzheimer's disease? |
| 60 | What are the assumptions of experiments measuring protein kinetics? |
| 61 | Why is autosomal dominant Alzheimer's disease used as a model for sporadic Alzheimer's disease? |
| 62 | How are anti-amyloid antibodies generated? |
| 63 | Why does hyperphosphorylation of tau cause it to aggregate? |
| 64 | How does vascular dysfunction affect Alzheimer's disease? |
| 65 | What regions of the brain show the earliest amyloid deposition? |
| 66 | What regions of the brain show the earliest tau accumulation? |
| 67 | What changes in tau are specific to Alzheimer's disease and not seen in other tauopathies? |
| 68 | Describe the process of neurofibrillary tangle formation? |
| 69 | How are tau neurites different from neurofibrillary tangles? |
| 70 | What is the best way to measure delay in disease onset with anti-amyloid treatments? |